\documentclass{article} 
\usepackage{collas2024_conference,times}
\usepackage{easyReview}
\usepackage{times}
\usepackage[utf8]{inputenc} 
\usepackage[T1]{fontenc}    
\usepackage{url}            
\usepackage{amsfonts,amsmath,amssymb}       
\usepackage{xcolor}
\usepackage{nicefrac}       
\usepackage{multirow}
\usepackage{algorithm}
\usepackage{algpseudocode}
\usepackage[normalem]{ulem}
\usepackage{arydshln}
\usepackage{caption}
\usepackage{wrapfig}
\usepackage{enumitem}
\usepackage{microtype}
\usepackage{graphicx}
\usepackage{subfigure}
\usepackage{amssymb}
\usepackage{booktabs} 
\usepackage{amsmath}
\usepackage{cutwin}
\usepackage[pangram]{blindtext}
\usepackage[calc]{adjustbox}
\usepackage{color}
\usepackage{authblk}
\usepackage{easyReview}
\usepackage{comment}
\usepackage{hyperref}       
\usepackage{booktabs}       
\usepackage{microtype}      
\usepackage{mathtools}
\usepackage{subcaption}
\usepackage{makecell}
\usepackage{amsmath}
\usepackage{amsmath,amsfonts,bm}

\def\eqref#1{equation~\ref{#1}}

\def\1{\bm{1}}

\DeclareMathAlphabet{\mathsfit}{\encodingdefault}{\sfdefault}{m}{sl}
\SetMathAlphabet{\mathsfit}{bold}{\encodingdefault}{\sfdefault}{bx}{n}

\usepackage{hyperref}
\hypersetup{
    colorlinks=true,
    linkcolor=red,
    filecolor=magenta,
    urlcolor=blue,
    citecolor=purple,
    pdftitle={Overleaf Example},
    pdfpagemode=FullScreen,
    }

\title{Masked Autoencoders are Efficient Continual \\ Federated Learners}

\author[*]{Subarnaduti Paul$^{1,2*}$, Lars-Joel Frey$^{1}$, Roshni Kamath$^{1,2}$, Kristian Kersting$^{1,2,3,4}$, Martin Mundt$^{1,2}$}
 
\affil[1]{Department of Computer Science, TU Darmstadt, Darmstadt, Germany}
\affil[2]{Hessian Center for AI (hessian.AI), Darmstadt, Germany}
\affil[3]{German Research Center for Artificial Intelligence (DFKI), Darmstadt, Germany}
\affil[4]{Centre for Cognitive Science, TU Darmstadt, Darmstadt, Germany}
\affil[*]{\{subarnaduti.paul, martin.mundt\}@tu-darmstadt.de}

\collasfinalcopy 

\begin{document}

\maketitle

\begin{abstract}
Machine learning is typically framed from a perspective of i.i.d., and more importantly, isolated data. In parts, federated learning lifts this assumption, as it sets out to solve the real-world challenge of collaboratively learning a shared model from data distributed across clients. However, motivated primarily by privacy and computational constraints, the fact that data may change, distributions drift, or even tasks advance individually on clients, is seldom taken into account. The field of continual learning addresses this separate challenge and first steps have recently been taken to leverage synergies in distributed settings of a purely supervised nature. Motivated by these prior works, we posit that such federated continual learning should be grounded in unsupervised learning of representations that are shared across clients; in the loose spirit of how humans can indirectly leverage others' experience without exposure to a specific task. For this purpose, we demonstrate that masked autoencoders for distribution estimation are particularly amenable to this setup. Specifically, their masking strategy can be seamlessly integrated with task attention mechanisms to enable selective knowledge transfer between clients. We empirically corroborate the latter statement through several continual federated scenarios on both image and binary datasets.
\end{abstract}

\section{Introduction}
Whether from a purely practical or from a biologically plausible perspective, building machines that mirror the capabilities of humans \citep{lake2017building} requires the ability to continue learning throughout their lifetime \citep{chen2018lifelong}. The key challenge is often framed as the sequential learning problem, where, much in contrast to traditional static train-validation-test dataset splits, neural models suffer from catastrophic interference \citep{McCloskey1989}. They tend to forget what they have previously seen if revisits are disallowed. When surveying the literature landscape, several biological underpinnings \citep{kudithipudi2022underpinnings} and practical pillars \citep{hadsell2020embracing, mundt2023} to alleviate the catastrophic forgetting phenomenon are typically at the focus of attention. Considerations span from regularization, dynamic architecture, and popular episodic memory buffer techniques \citep{hayes2021rehearsal} to tangible quantities to measure compute, parameter growth, and various other performance assessments surrounding knowledge transfer and accumulation over time \citep{mundt2022cleva}.  

\begin{figure}[t]
\centering
\includegraphics[width=0.8\textwidth] {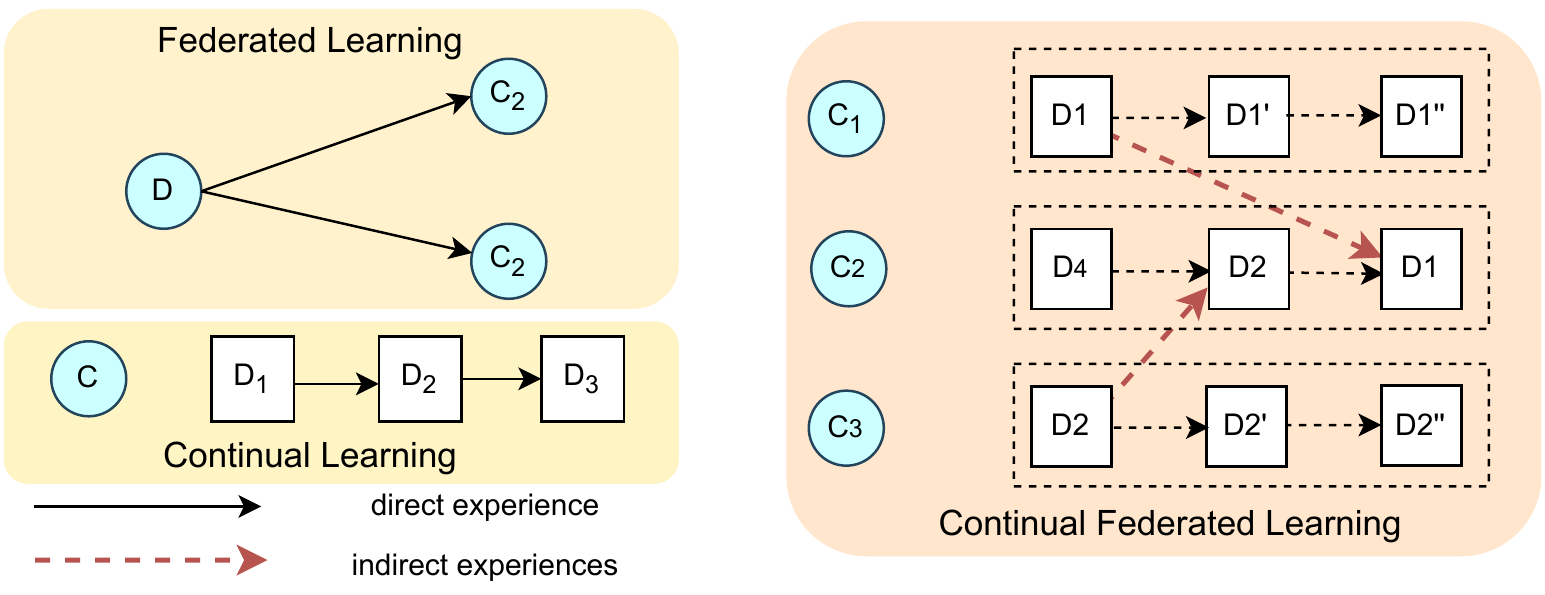} 
\caption{Set-up schematic. As data (on clients) may drift individually over time, models need to mitigate forgetting. In distributed scenarios, being additionally informed through indirect experience (dashed red arrows) provides further learning benefit, as other clients may observe similar data at different points in time.  
\label{fig:fedcl}}
\end{figure}

However, the primary objective of continual machine learning (CL) seems to be predominantly centered around maintaining a single model, heavily inspired by an individual human's ability to learn throughout their lifetime. At the same time, well-known research from social and cognitive sciences suggests that humans also heavily benefit not only from recalling their own experiences but also leveraging the knowledge of their fellow humans. In other words, they learn from indirect experiences. Take for instance an organization working together, where every member partially profits from the growing know-how simply by listening to the report of others \citep{argote2011organizationallearning, gino2010indirectexp}. In addition to each individual's episodic memory, there thus exists the notion of a transactive memory \citep{wegner1985, wegner1987}. Relating back to machine learning, an intuitive parallel can be found in the concept of federated learning (FL) \citep{FedAvg, kairouz21fedadvances}, where several learning clients attempt to share and consolidate their knowledge, most commonly with a central server. Alas, FL comes with its own frequent focus, motivated primarily by computational efficiency and preservation of privacy by avoiding the explicit communication of data. As such, a recent survey \citep{criado2022longroad} argues that FL has, quoting the original authors, a ''long road ahead``. The latter is attributed to the fact that while FL seems to distribute data and minimize communication, it seldom considers scenarios where data distributions on clients drift or individual tasks change over time. Speaking informally, although continual learning and federated learning share similar properties in data being distributed across ''time`` or ''space`` respectively, their joint consideration remains largely open. We point to figure \ref{fig:fedcl} for visual intuition. 

Thus, inspired by the notion of transactive memory and the necessity to combine federated and continual learning, we propose an unsupervised continual federated learner (CFL), focused on selectively sharing acquired representations. More specifically, we build on the previous work of federated weighted inter-client transfer (FedWeIT) \citep{FedWeit}, which has proposed an attention based approach to CFL for supervised classification scenarios. In particular, we leverage their proposed parameter decomposition to intuitively integrate it with a masked autoencoder for distribution estimation (MADE) \citep{MADE}. We demonstrate that MADE, an effective distribution estimator,  is particularly amenable to unsupervised CFL, due to its inherent auto-regressive masking strategy sharing a common denominator with attention based masking to avoid forgetting. In summary, our specific contributions are: 
\begin{itemize}
    \item We draw inspiration from the supervised FedWeIT and extend it to our unsupervised \textbf{Con}tinual \textbf{Fed}erated \textbf{MA}sked autoencoders for \textbf{D}ensity \textbf{E}stimation (CONFEDMADE); an unsupervised continual federated learner based on masking to enable selective knowledge transfer between clients and reduce forgetting. 
    \item We highlight that MADE is a model particularly amenable to unsupervised CFL and investigate several non-trivial considerations, such as connectivity and masking strategy, beyond a trivial application of federated averaging and FedWeIT to the unsupervised setting. 
    \item We extensively evaluate our approach on several CFL scenarios on both image and numerical data. Overall, CONFEDMADE consistently reduces forgetting while sparsifying parameters and reducing communication costs compared to other unsupervised CFL approaches.
\end{itemize}

\section{Related Work}
We briefly introduce the key principles behind typical federated and continual learning set-ups, as well as their conjunction into CFL, illustrated in figure \ref{fig:fedcl}. In the process, we provide a concise overview of existing approaches, from a perspective of unsupervised learning to motivate our proposed CONFEDMADE. 


\textbf{Federated Learning:}  
A standard FL set-up collaboratively trains $C$ number of clients $\{c_1, c_2, \ldots, c_C \} $, typically aggregated to yield a global server model \citep{kairouz21fedadvances}. Here, each client learns on its privately accessible dataset, with the key objective to keep this data private without sharing it explicitly. Instead, the pioneering and to date most popular approach is to communicate parameters or gradient signals and average them: so called federated averaging (FedAvg) \citep{FedAvg}. In practice, the focus of FL seems to generally lie on preserving privacy while lowering communication and computation costs. Whereas data heterogeneity becomes a challenge, it is predominantly the case because a single task $t$ comprised of dataset $D = \{ \boldsymbol{x_1}, \boldsymbol{x_2}, \ldots, \boldsymbol{x_n}\}$ is distributed across the $C$ clients. As each client only sees a fraction $N/C$ of the data, it can no longer be assumed that data is i.i.d. overall. 

Different FL algorithms have been proposed to deal with this challenge. For instance, FedProx \citep{FedProx} introduces an additional proximity term to deal with deviations in data distribution across clients. FedBN \citep{FEDBN} improves convergence speed by leveraging batch normalization layers, in a similar spirit to various works that have focused primarily on computational reduction \citep{FedMD, FEDBN, FEDGMS}. Analogously, select works \citep{FRecon, MOON} aim to reduce computation and memory overhead, e.g.~through reconstructing local parameters or formulating contrastive losses. Most of these approaches could intuitively find application in unsupervised approaches, where far fewer works have formulated explicitly tailored mechanisms. Here, dedicated techniques often frame the problem from a clustering perspective 
\citep{lubana2022orchestra, lu2022federated, chung2022federated, zhang2020fedusupcontr}, reducing communication overheads in the process. Other works once more employ contrastive techniques to align representations through distillation \citep{han2022fedx} or by fragmenting training into multiple stages \citep{berlo2020fedrep}.
However, none of these works consider the case where the data distribution for each client in the distributed setup may severely shift over time, or where client-specific tasks may change independently. In other words, although data heterogeneity is a prominent theme, continual learning seldom is.

\textbf{Continual Learning:} In continual learning, data heterogeneity is typically addressed from a different, complimentary perspective to the FL set-up. A standard CL setup trains a single (client) model on a sequence of $\{ \mathcal{T}^{(1)}, \mathcal{T}^{(2)}, \ldots, \mathcal{T}^{(T)}$\} tasks. This can be expressed through observing ${\boldsymbol{x}^{(t)}_n}$ of $N^{(t)}$ individual instances in a $t^{th}$ task's dataset $D^{(t)}$, where generally $p(\boldsymbol{x}^{(t)}) \neq p(\boldsymbol{x}^{(t+1)})$. The earlier mentioned surveys in the introduction \citep{chen2018lifelong, kudithipudi2022underpinnings, hadsell2020embracing, mundt2023} highlight how the focus is then typically on designing mechanisms to avoid catastrophic forgetting in this single model setup, that predominantly falls into three general categories of regularization (e.g. elastic weight consolidation (EWC) \citep{EWC}, rehearsal (e.g. gradient episodic memory \citep{GEM}), or dynamic architectures (e.g. dynamically expandable architectures \citep{DEN}).  
Unsupervised continual learning approaches can correspondingly also be attributed to any of these three main pillars of CL. Apart from works on lifelong generative adversarial networks \citep{Goodfellow2014, ramapuram2020lifelong}, autoencoders \citep{Ballard1987ModularLI} trained with variational inference \citep{Kingma2013} present a frequent choice. Both are popular due to their ability to rehearse previously observed examples through a generative model \citep{Shin2017}. Several unsupervised CL follow-ups \citep{Achille2018, rao2019continual, Mundt2022unified} additionally propose how latent space structuring techniques aid in removing assumptions that have resulted in the majority of CL algorithms being tailored to incremental classifiers \citep{Farquhar2018, mundt2022cleva}, shifting towards considerations on data distributions. However, unifying unsupervised CL with an additional challenge of data being distributed  across multiple clients with changing tasks is yet to be considered by these works.

\textbf{Continual Federated Learning:} 
Formally, a continual federated learner will contain C clients $\{c_1 , c_2  , \ldots, c_C \}$, where each client will individually observe a sequence of tasks and each individual task $\mathcal{T}^{(t)}_{c}$ consists of a data subset $D^{t}_{c} = \{ \boldsymbol{x_1}, \boldsymbol{x_2}, \ldots, \boldsymbol{x_n}\}^{t}_{c}$. A single task for a single client thus contains $N^{(t)}_c$ individual instances ${\boldsymbol{x}^{(t)}_{c,n}}$ that are trained on for overall $e$ epochs with $r$ communication rounds to the outside. Following figure \ref{fig:fedcl}, a client could observe a task that a different client has observed at a previous time step, i.e.~observing instances from the same distribution a separate client has already seen at a different time step, or alternatively all data seen by clients at all times can be distinct. \\ 
Intuitively, one could attempt to apply any of the three pillars of CL to this distributed scenario, as algorithms such as FedAvg can directly be applied to any unsupervised learner. In fact, \citep{park2021fedinc} follows this idea and couples training with a rehearsal strategy, whereas the supervised \citep{usmanova2021confeddistill} proposes the use of distillation. However, the latter is tailored to supervised scenarios and the former seems to be challenging to integrate into the typical federated learning desideratum of protecting privacy (by employing an explicit memory buffer). Similarly, we could conjecture the use of the variational CL procedures outlined in the former subsection, but note that they do not easily allow for explicit data distribution estimates. Conceptually, our work thus follows a different approach, referred to as federated weighted inter-client transfer (FedWeIT) \citep{FedWeit}. FedWeIT is a supervised framework that is cohesive with early postulated principles of modularity through reduction of parameter overlap \citep{French1992}. Specifically, it decomposes parameters into generic and task-specific ones and proceeds to learn attention maps to share and preserve appropriate subsets. In our work, we propose a synergy between two independent masking strategies and show that this prior art of FedWeIT can be effectively integrated with an efficient distribution estimator based on masked autoencoders (MADE) \citep{MADE} and FedAvg into an unsupervised cost-effective CFL approach. 

\section{Unsupervised Continual Federated Learning with Masked Autoencoders}
We now formally introduce our unsupervised CFL approach with the aim for a client to learn meaningful representations continually throughout their lifetime while leveraging other clients' indirect experiences. To mirror realistic real-world set-ups, the summarized key elements of CFL are 1.) each client holds its sequence of privately accessible task data; 2) the distribution of tasks across the clients may be different; 3) depicting a common continual learning assumption, a particular client is practically unaware of the ordering of the set of tasks and a particular set of data is only seen once in each client's lifetime. Before introducing why masked auto-encoders are suitable to operate in CFL scenarios to learn meaningful representations, we briefly introduce required preliminaries and analyze important considerations when embedding masked autoencoders to a federated scenario (without the integration of an extra continual element). 
%
\subsection{Preliminaries: Federated Weighted Inter-client Transfer \& Masked Autoencoding}
\textbf{FedWeIT} \citep{FedWeit} has been proposed to tackle data heterogeneity in supervised CFL scenarios. Here, with data distribution varied across clients, a traditional FedAvg \citep{FedAvg} will suffer from negative interference from other clients, typically leading to the global model averaging an improper solution. Unlike traditional CL strategies in earlier surveys, FedWeIT decomposes parameters and selectively transmits previously acquired knowledge across the client network. The trainable client model parameter $W_c^{t}$ at task t is first additively decomposed into two sets of parameters: one assigned to capture its generic knowledge known as local-base parameter $B_c^{t}$ and another assigned to capture its task-specific knowledge $A_c^{t}$ for each individual task $t$. The Server in turn stores the learned task-adaptive parameters into a ``knowledge base'' $K_b$ which is then accessible for future training steps. To selectively utilize previously learned experiences, each client therefore learns to allocate a weighted (scalar) task-specific attention  $\alpha^{t}$ for all other clients' $C/c$ past tasks $j<|t|$:
\begin{equation}
  \begin{aligned}
      W_{c}^{t} = B_{c}^{t}  * m_{c}^{t} + A_{c}^{t} + \sum_{i\epsilon C_{/c}} \sum_{ j< |t|} \alpha_{(i,j)}^{t} * A_{i}^{(j)} \quad 
    \label{eq:fedweit}  
  \end{aligned}   
\end{equation}
In optimization, a threshold mask $m_c^{t}$ is then learned to additionally reduce inter-client interference:

\begin{equation}
   \begin{aligned}
    \underset{B^{(t)}_c, A^{(1:t)}_c,\alpha^{(t)}_c,m^{(t)}_c}{min} \mathcal{L} (W^t_c, T^{(t)}_c) +
    \lambda_1 \Omega (\{ m^{(t)}_c,A^{(1:t)}_c\})+ \lambda_2 \sum_{i=1}^{t-1} \parallel \Delta B^{(t)}_c \odot m^{(i)}_c + \Delta A^{(i)}_c  \parallel^2_2 
\end{aligned} 
\label{eq:fedweitloss}
\end{equation}
Here, the first term $\mathcal{L}$ is a regular cross-entropy loss for each client $c$'s individual task  $\mathcal{T}^{(t)}_{c}$.The first additional term $\Omega$ is a sparsity inducing regularization term (typically chosen as L1) to ensure $A_c^{t}$ is highly task-specific, whereas a second (L2) regularizer is used to reduce catastrophic forgetting with respect to prior time steps. The latter is achieved by limiting the change in local-base parameters $\Delta B_c^{(t)}$ and the change in task-adaptive parameters $\Delta A_c$ between time steps. $\lambda_1$ and $\lambda_2$ are hyperparameters to weigh the individual terms.

\textbf{Masked Autoencoders} \citep{MADE} have been originally proposed for the purpose of distribution estimation (MADE). We will later highlight why they are particularly suitable for CFL beyond their demonstrated capability for unsupervised learning, after delving into some important considerations and caveats. In particular, MADE adheres to the auto-regressive property through the use of masking, where an output unit $\hat{x}_d$ only depends on previous input units $x_{<d}$, assuming a d-dimensional input. Hence, the autoencoder outputs take the form $\hat{x}_d = p(x_d | x_{<d})$, where $p(x_d)$ is the probability of observing $x_d$ at $d^{th}$ dimension, serving a dual purpose of a data distribution estimator as well as a synthetic data generator. In the below equation \ref{eq:MADE}, we compute the output {$h^l(x)$} for hidden layers $l = 1, 2, \ldots, L$ and the reconstructed input $\hat{x}$ of MADE:
\begin{equation}
\begin{aligned}
h^l(x) = g(b^l+(W^l\odot M^{W^l})h^{l-1}(x))\\ 
\hat{x} = \text{sigm}(c + (V \odot M^V)h^L(x) + (D \odot M^D)x)
\label{eq:MADE}
\end{aligned}
\end{equation}

\begin{figure*}[t]
\begin{center}
\includegraphics[width=1.0\textwidth]{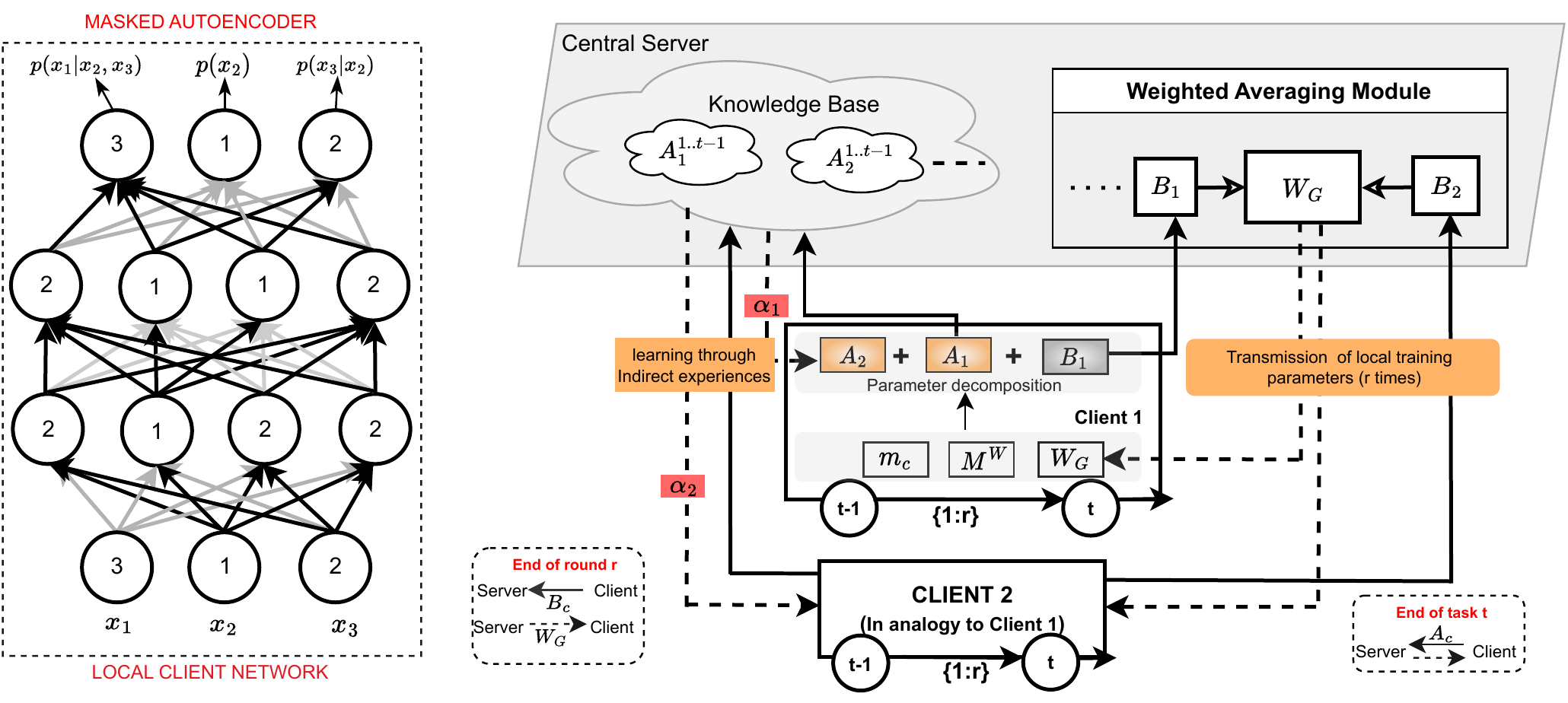}
\caption{ Proposed CONFEDMADE framework. On the left, we show a visual representation of a masked autoencoder architecture based on the MADE masking rule in eq. 4. The central server, on the right, initializes the global model and distributes it to all the participating clients along with the masking variables $M^W$ and federated mask $m_c$. In the client module, masked model parameters ($W_G * M^W * m_c$) refers to the local client network (left figure) which are then additively decomposed into task-specific $A_c$ and local-base parameters $B_c$ to obtain the final set of trainable parameters. After each round of training $r$, clients communicate local-base parameters (in dotted lines) to the server whereas it only communicates task-adaptive parameters after completing r rounds of training (in solid lines). The central server computes the averaged weighted parameter via Fed-Avg and communicates it back to the clients at the beginning of each round, whereas it stores $A_c$ into the knowledge base and communicates only at the beginning of each new task $t$.}
\label{fig:FCMADE_framework}
\end{center}
\end{figure*}

Here, $\hat{x}$ is the reconstructed output for a given data input $x$,  "sigm" the sigmoid function, and $"g"$ an arbitrary activation function. $b$ is a layer's bias, whereas $W, V, D$ denote the connection matrices for input to hidden layer, hidden to output layer, and a direct (skip) connection (DC) between the input and output layer respectively. The respective binary mask matrices $ M^W, M^V$ for hidden layer, and output layer connections respectively are defined as (D defined analogously):

\begin{minipage}{.5\linewidth}
\begin{equation*}
    \begin{aligned}
        M^W_{k,d}= 1_{m(k)\geq d} = 
        \begin{cases}
            1 
             & \text {if m(k) $\geq$ d}
        \\[4pt]
        0 
        & \text{otherwise,}
        \end{cases}
\label{eq:binary_mask_1}
\end{aligned}
\end{equation*}
\end{minipage}%
\begin{minipage}{.5\linewidth}
\begin{equation}
    \begin{aligned}
            M^V_{k,d}= 1_{d > m(k)} = 
        \begin{cases}
            1 
             & \text {if d $>$ m(k)}
        \\[4pt]
        0 
        & \text{otherwise,}
        \end{cases}
\end{aligned}
\label{eq:MADE_mask_equation}
\end{equation}
\end{minipage} 

where k $\in \{1..K\}$ refers to hidden layer units. Following equation \ref{eq:MADE_mask_equation}, MADE assigns binary masks as we determine $m(k)$ for the k-th unit by sampling random integers less than D to the hidden units. 
Instead of learning a single deterministic model, we can further opt for \textbf{order-agnostic training} (OA) (resampling input ordering after each mini-batch, thereby reproducing the mask for the first hidden layer) or \textbf{connectivity-agnostic training} (CA) (resampling the hidden layers units). 
We point to the left side of figure \ref{fig:FCMADE_framework} for an illustration of a MADE network and proceed to exploit the masking in MADE as a part of our CFL framework, thereby significantly reducing communication costs. 
%
\subsection{Embedding a Masked Autoencoder into the Federated-Averaging Framework}
Before establishing MADE as an efficient continual federated learner, we first analyze a traditional FL setup. Hence, we first take a step back from CL and distribute data from the same distribution across multiple clients, and perform 

\begin{minipage}[t!]{\textwidth}
  \begin{minipage}[t]{0.485\textwidth}
    \captionof{table}{Performance of MADE (\textit{NLL Loss}) with synchronized \& distinct masks for federated MNIST. In the former, random masks (or the respective seed) gets communicated to all clients from the server, in the latter masks are drawn independently.}
    \label{tab:FEDMADE_mask}
    \resizebox{\linewidth}{!}{%
    \begin{tabular}{@{}lcr@{}}
    \toprule
     Clients & Synchronized Mask & Distinct Mask \\
    \midrule
     1   & 75.23 $\pm$ 0.27   &   75.23 $\pm$ 0.27 \\ 
     2   & 76.93 $\pm$ 0.35    &  97.27 $\pm$ 0.21 \\
     5   & 83.51 $\pm$ 0.83    &  170.7 $\pm$ 0.66 \\
     10   & 86.23 $\pm$ 0.99    &  219.1 $\pm$ 1.54 \\
     20   & 87.89 $\pm$ 1.11    &  264.0 $\pm$ 3.57 \\
     40  & 89.89 $\pm$ 1.47    &  301.2 $\pm$ 2.15 \\
    \end{tabular}%
    }
    \label{tab:syn}
  \end{minipage}
  \hfill
  \begin{minipage}[t]{0.485\textwidth}
   \captionof{table}{Performance of Federated MADE (\textit{NLL Loss}) for different architectural choices, including direct connections, connectivity- and order-agnostic training, as well as their combinations. \label{tab:FEDMADE_choices} \smallskip}
  
       \resizebox{\linewidth}{!}{%
    \begin{tabular}{@{}lcc@{}}
    \toprule
    MADE configuration & Client & Server \\
    \midrule
    Baseline  & 89.66 $\pm$ 0.61  & 100.8 $\pm$ 0.66 \\
    \midrule
    +Direct Connection & 75.66 $\pm$ 0.57  & 84.10 $\pm$ 0.59 \\
    
    +Connectivity Agnostic & 98.32 $\pm$ 0.22  & 110.7 $\pm$ 0.60 \\
    +Order Agnostic & 81.18 $\pm$ 0.64  & 84.78 $\pm$ 2.37 \\
    \midrule

    +Direct+Order & 62.83 $\pm$ 0.30  & 71.42 $\pm$ 1.19 \\
    +Direct+Connectivity & 78.64 $\pm$ 1.74  & 87.12 $\pm$ 1.76 \\
    \midrule
    +DC+CA+OA & 103.6 $\pm$ 0.26  & 111.5 $\pm$ 0.97 \\
    \end{tabular}%
    }
        \smallskip
       
      \label{tab:comm}
    \end{minipage}
\end{minipage} \\

FedAvg with the help of a central server. For each client, we assign a MADE model based on the principles discussed in the previous section. In particular, we wish to investigate how a federated MADE (FEDMADE) performs when i) an increasing amount of clients get a smaller percentage of data, ii) FEDMADE is confronted with only a (skewed) subset of data, to anticipate the later continual learning set-ups, iii) and most importantly, whether and how the masking in different clients may interfere with each other, iv) Finally, we consider different training choices such as OA, CA, etc and their combinations and show how such modifications impact the learning efficacy of the federated MADE.

To analyze and answer these questions, we report the performance in terms of obtained negative log-likelihoods, where smaller numbers are better, for an unsupervised federated MNIST \citep{MNi} set-up. Each of the $C$ clients receives $1/C^{th}$ of the data without any labels. In table \ref{tab:FEDMADE_mask}, we answer the above questions i) and iii) simultaneously and highlight the most important aspect of moving MADE into a federated scenario. Namely, we observe that sampling random integers for the masks, as suggested in the original MADE framework, independently on each client network, is detrimental to the obtained performance. The rationale is quite intuitive, as data gets distributed across an increasing amount of clients, averaging different MADE clients presently update different sub-models at each communication round. As a crucial step to penable effective FEDMADE and thus also later continual FEDMADE, we are required to communicate a similar sequence of mask(s) from the server to all the participating clients. Although masks are small, this in fact incurs an extra communication expense, which can however be almost fully alleviated if the different clients operate on similar devices and operating systems that can interpret the same random seed in the same manner, in which case communicating the latter is sufficient. The respective "synchronized mask" column in table \ref{tab:FEDMADE_mask} demonstrates how this almost fully alleviates the performance drop, even when a large number of clients is employed in a FL network. \\

Table \ref{tab:FEDMADE_choices} complements this insight with answers to the remaining two questions, how different MADE choices affect performance, and how a skewed subset of data affects learned representations. For this purpose, we emulate the first step of a CL scenario and train a federated MADE on distributed MNIST with three classes. We can observe how this results in a perhaps expected loss in performance of the baseline without any of the optional architectural configurations. Naively, one would however equip MADE with all components that have initially been hypothesized to yield advantages in its original paper. Here, we observe a pattern that is consistent with our previous observation on masking. In fact, direct connections provide a cohesive benefit, but this is not necessarily the case for order and connectivity-agnostic training. The former's resampled input ordering after each mini-batch independently on clients, as posited in the original work, leads to more robust models. The latter's resampling on the level of hidden units however massively degrades performance due to interference with FedAvg, following the same intuition as with the earlier independent masking. As these considerations are imperative, we take them into account in our ensuing design of continual federated MADE. 

\subsection{CONFEDMADE: Learning Representations Continually through Indirect Experience}
In the previous section, we laid out the foundation to employ MADE in a federated scenario. A critical next step is to inherit a form of continual learning strategy that can handle catastrophic forgetting under data distribution shifts. The most direct way would be to fully integrate MADE into the earlier outlined FedWeIT framework. We will refer to this strategy as FedWeIT-MADE. 
However, we can move beyond such a combination. To this end, we propose CONFEDMADE, which on the one hand includes our earlier findings of section 3.2 (federated MADE) and on the other hand, incorporates its masking strategy based on the auto-regressive property in a more meaningful way into continual learning. This will outperform the naive combination while simultaneously cutting down the communication cost. 

Figure \ref{fig:FCMADE_framework} provides a conceptual overview of our proposed CONFEDMADE framework. First, note that the server serves a dual purpose: 1) averaging the learned representations of the individual clients and communicating global weights $W_G$ back along with the auto-regressive mask  $M^W$ and federated mask $m_c$ (to select the relevant parameters); 2) building a memory buffer as knowledge base for storing the task-specific representations learned by all the clients in previous time steps.
Second, we also decompose the parameters of the clients' MADE models, fusing earlier equations \ref{eq:MADE} and \ref{eq:fedweit}. Thus, the overall trainable model parameters {$W_C$} become an auto-regressivly masked $M^W$ variant:.   
\begin{equation}
\begin{aligned}
    W_{C} \odot M^W = B_{c} \odot m_{c} \odot  M^W + A_{c} \odot M^W + 
    \sum_{i\epsilon C_{/c}} \sum_{ j< |t|} \alpha_{(i,j)} * A_{i}^{(j)} \odot M^W \quad 
\end{aligned}
\label{eq:Confed}
\end{equation}
Unlike the supervised FedWeIT, this now ensures ensure that an identical auto-regressive masking strategy is imposed on each of our clients in the federated unsupervised network, consistent with our earlier findings of section 3.2. Finally, our proposed CONFEDMADE approach results in two masks being applied to the local-base parameter $B_c$, that despite masking the same set of weights, would be treated independently. We could then treat masked autoencoders as our choice of client architecture and optimize using earlier equation \ref{eq:fedweitloss}.  

However, once more, this is not only practically wasteful by not exploiting synergies, but can also hurt performance through interference effects when employing FedWeIT's regularizers to pull towards a set of parameters that could presently be masked differently by the autoencoder. To both combat this interference of two independent masks (to clarify, for task adaptive parameters and for MADE) and make use of the sparsity induced by the MADE mask at any point in time, we thus optimize the decomposed MADE client parameters through a modified loss function:
\begin{equation}
\begin{aligned}
    \underset{{B_c^{(t)}, A_c^{(1:t)},\alpha^{(t)}, m_c^{(t)}}}{min} \mathcal{L} (W^t_{c}, \mathcal{T}^{(t)}_c) + \lambda_1(\Omega (\{ m^{(t)}_c, A^{(1:t)}_c \odot M^W \}))
     +\lambda_2 \sum_{i=1}^{t-1} \parallel \Delta B^{(t)}_c \odot m^{(i)}_c + \Delta A^{(i)}_c  \parallel^2_2 
\end{aligned}
\label{eq:Confed_loss}
\end{equation}
In this augmented variant of equation \ref{eq:fedweitloss} for our continual federated MADE, we thus synergize model parameter sparsification through masking while constraining optimization to only parameters that are presently meaningful (i.e. masked). 
Primarily, we have modified the sparsity-inducing term $\Omega$, which in practice is an L1 norm \citep{FedWeit}, to yield as small as possible amount of task-adaptive ($A_c$) parameters to populate the knowledge base. In practice, at any point in time, we include a client's present MADE mask $M^W$ in an element-wise product with the $A_c$ parameters. In this way, we avoid regularizing  parameters that are currently not contributing to learning the input distribution and thus also explicitly further encourage task-adaptive parameters to be even sparser, subject to the empirically chosen masking ratio of the masked autoencoder architecture. We analyse this trade-off in the later experimental section. 

The second, perhaps implicit modification, is to \textit{refrain} from adding the mask to the L2 regularizer, as would be intuitive according to earlier equation \ref{eq:Confed}, where the MADE mask operates on the local base parameters in conjunction with the FedWeIT mask. The reasoning here is two-fold. First, the sum operates over previous time steps, and as such a current steps' MADE mask would mask a different portion of parameters and thus interfere. Naturally, this could be solved by also storing respective MADE masks in the knowledge base, at the expense of larger memory and communication overheads. However, and interestingly, one can realize that we also do not need the MADE mask in this term. On the one hand, the $A_c$ parameters up to $t-1$ have already been largely sparsified through the $\Omega$-term. On the other hand, before computing the actual L2 term between the base and task-adaptive parameters,a difference in the level of $B_c$ and $A_c$ is taken individually first ($\Delta B_c$ \& $\Delta A_c$). By definition, masking a value to be zero does not contribute to differences or sums and vice-versa if a respective weight had indeed been masked and received no update, its difference between time steps would be zero independently of its exact value anyhow. Overall, the general procedure of CONFEDMADE can be summarized in the subsequent algorithm and involves the following communication cost:
\begin{minipage}[h!]{\textwidth}
    \begin{minipage}[b]{0.55\textwidth}
        \begin{algorithm}[H]
            \caption{\textbf{CONFEDMADE}: masked auto-encoder clients undergo parameter decomposition into base and task-adaptive parameters with the latter being stored in a server's knowledge base. The optimization procedure sparsifies (through regularization) parameters and clients learn to attend to relevant server knowledge while minimizing interference between clients through masking}\label{euclid}
            \hspace*{\algorithmicindent} \textbf{Input} number of clients $C$, datasets $\{D_c^{(1:t)}\}_{c=1}^C$ , global parameters $W_G$, hyperparameters $\lambda_1, \lambda_2$, $\lambda_3$, 
             knowledge base $kb = \{\}$\\
            \hspace*{\algorithmicindent} \textbf{Output} $\{B_{c}, m_{c}^{(1:t)}, \alpha_{c}^{(1:t)}, A_{c}^{(1:t)}\}_{c \in C}$
            \begin{algorithmic}[1]
            \State Initialize Sever Model with global parameter $W_G$
            
            \For {task $t = 1,2, ...$}
            
            \State Initialize $B_{c}$ to $W_G \, \forall c \in C$
            \State Initialize task-adaptive parameters $A_c$ to $B_c/\lambda_3 \, \forall c \in C$
            \For {round $r = 1,2, ...$}
            \State Define auto-regressive mask $M^W$ for MADE
            \State Distribute $W_G$, $M^W$, $kb^{t}$ to all clients $c \in C^{(r)}$ 
            \State Initialize $B_{c}$ to $W_G$ for all clients $c \in C$
            
            \State Decompose trainable parameters using equation 5
            \State Optimize loss terms in equation 6.
            \State Transmit $\{B_{c}^{(t,r)} \odot m_{c}^{(t,r)} \odot M^W\}$ from $C^{(r)}$ to server
            \State Fed. Avg: $W_G \gets \frac{1}{|C^{(r)}|} \sum_c B_{c}^{(t,r)}\odot m_{c}^{(t,r)} \odot M^W$
            \EndFor
            \State Communicate learned task-adaptive parameters $A_c$ back to the server.
            \State Update knowledge base $\{kb \gets kb \cup \{A_{j}^{(t)}\}_{j \in C}\}$
            \EndFor
            \end{algorithmic}
            \label{alg:ConFedMADE}
        \end{algorithm}
    \end{minipage}
    \hfill
    \begin{minipage}[b]{0.425\textwidth}
        \textbf{Cost of communication at end of round r:} The participating clients communicate the local base parameters to the server which then sends back the averaged global parameter. The communication cost for each client accounts to ($\Tilde{B_c} * |r|$), where $|r|$ is the total number of communication rounds (linked to SGD steps) and $\Tilde{B_c}$ is a masked subset of the local-based parameter $(|B_c \odot m_c \odot M^W|)$ for each client. In reference to FedWeIT \citep{FedWeit}, the amount of communicated (non-zero) parameters is thus reduced through additional auto-regressive mask in the form of MADE mask. The communication cost for a central Server equates to ($\Bar{W_G} * |r| * |C|$), where $\Bar{W_G}$ denotes the global parameter obtained through averaging local base parameters of each clients in the network, with a total of $|C|$ number of clients. \\
        
        \textbf{Cost of communication at end of round t:} After completion of r ``inner'' communication rounds, clients communicate the learned task-specific representations $A_c$ which is stored in a separate memory maintained by the central server. The Server then communicates the stored $A_c$ from previous tasks to the clients at the beginning of each new task. The total communication cost accounts to $|C| *  (|R|*|W_G|+A_c)$. For reference, a communication overhead of a typical FL approach is  $|C| *  (|R|* \theta)$ where $\theta$ is the full set of model parameters without decomposition or a subset masking.
    \end{minipage}
\end{minipage}
%
%
\section{Experimental Analysis}
We empirically corroborate the advantages of CONFEDMADE in three different dataset sequences while contrasting it with several baseline approaches. Specifically, we quantify how i) forgetting is mitigated, ii) adaptive task knowledge is attended to reduce inter-client interference, and iii) communication costs are reduced across the network.
\subsection{Learning settings and compared approaches}
\textbf{Dataset sequences.} We construct three sequences of unsupervised tasks, ranging from images to numerical data.
\textit{A) Sequential MNIST:\citep{MNi}} comprised of 10 distinct digit classes, we consider 5 consecutive tasks per client to each consist of images with a single random digit. \textit{B) Numerical binary data:} we create sequences across four popular binary datasets from distribution estimation literature, namely ``RCV1'', ``Adult'', ``Connect4'', and ``Tretail'' \citep{binaryref1, UCI} with 100000 data instances in total. At each time step, a client is trained on only one of these numerical datasets. \textit{C) MNIST + EMNIST:} \citep{MNi,emnist} With a total of 36 distinct subsets, 10 digits and 26 letters, and a total of 215,600 data samples, a client observes mutually exclusive samples from either digits or letters in the form of a task. \emph{We specifically emphasize that we refer to labels only for the sake of intuition in formulating sequences of tasks for each clients (absolute data heterogeneity) and not for the fully unsupervised optimization (no task conditioning, labels, etc.).}  

\begin{table*}[t]
  \caption{Average final negative log-likelihoods and forgetting values (lower is better) for different learning settings and models. CONFEDMADE improves substantially through the symbiosis of auto-regressive masking and parameter decomposition, approaching the upper-bound more closely than an unsupervised FedWeIT.}
  \resizebox{\linewidth}{!}{
  \begin{tabular}{p{4cm}p{0.4cm}p{0.4cm}|lr|lr|lr}
       \multicolumn{3}{c|}{ }  & \multicolumn{2}{c|}{\textbf{MNIST}} &\multicolumn{2}{c|}{\textbf{Binary}} & \multicolumn{2}{c}{\textbf{MNIST, EMNIST}}\\
        \textbf{Learning Setting} & \textbf{FL} & \textbf{CL} & \textbf{NLL} ($\downarrow$) & \textbf{ Forgetting} ($\downarrow$) & \textbf{NLL} ($\downarrow$) & \textbf{ Forgetting} ($\downarrow$) & \textbf{NLL} ($\downarrow$) & \textbf{ Forgetting} ($\downarrow$) \\  
    \midrule
    Offline  & $\times$  &  $\times$  & 72.68 $\pm$ 1.68 & - & 38.45 $\pm$ 1.33 &- & 84.98 $\pm$ 1.21 & - \\ 
    
    Federated Offline &$\checkmark$ & $\times$  & 79.23 $\pm$ 1.11 & - & 40.33 $\pm$ 0.88 & - & 89.66 $\pm$ 3.12 & - \\
    \midrule
    CL-Cumulative Replay &  $\times$ &$\checkmark$ & 73.32 $\pm$ 1.23& 0.00 $\pm$ 0.00 & 39.45 $\pm$ 0.37 & 0.00 $\pm$ 0.00 & 86.66 $\pm$ 3.12 & 0.00 $\pm$ 0.00 \\ 
    EWC  & $\times$  &  $\times$  & 111.2 $\pm$ 0.86 & 29.33 $\pm$ 0.12 & 79.80 $\pm$ 0.35 & 26.01 $\pm$ 0.44 & 121.03 $\pm$ 0.89 & 27.87 $\pm$ 0.32 \\ 
    CL-Finetune  &  $\times$ & $\checkmark$& 126.1 $\pm$ 3.32 & 38.62 $\pm$ 2.76 & 81.32 $\pm$ 0.97 & 28.32 $\pm$ 1.23 & 125.3 $\pm$ 2.76 & 30.98 $\pm$ 0.43 \\ 
    \midrule
    FedCL-Cumulative Replay  &$\checkmark$ &$\checkmark$ & 74.47 $\pm$ 0.57 & 0.00 $\pm$ 0.00  & 41.67 $\pm$ 1.26 & 0.00 $\pm$ 0.00 & 87.34 $\pm$ 2.24 & 0.00 $\pm$ 0.00 \\ 
    FedProx &$\checkmark$ &$\checkmark$ & 106.34 $\pm$ 0.12 & 24.35 $\pm$ 0.67 & 73.34 $\pm$ 0.20 & 20.95 $\pm$ 0.27 & 117.67 $\pm$ 0.12 & 22.29 $\pm$ 0.32 \\ 
    FedCurv   &$\checkmark$ &$\checkmark$ & 104.94 $\pm$ 0.56 & 22.95 $\pm$ 1.13 &  70.34 $\pm$ 0.66 & 19.12 $\pm$ 0.33 & 116.11 $\pm$ 0.65 & 21.56 $\pm$ 0.43  \\ 
    FedProx + EWC &$\checkmark$ &$\checkmark$ & 105.97 $\pm$ 0.78 & 23.89 $\pm$ 0.78 & 73.04 $\pm$ 0.13 & 20.35 $\pm$ 0.23 & 116.94 $\pm$ 0.23 & 21.98 $\pm$ 0.32\\ 
    
    FedCL-Finetune  & $\checkmark$ & $\checkmark$ & 115.3 $\pm$ 5.67 & 31.43 $\pm$ 1.21 & 84.45 $\pm$ 0.45 & 29.56 $\pm$ 2.54 & 129.8 $\pm$ 3.21 & 36.98 $\pm$ 1.02 \\ 
    \midrule
    FedWeIT-MADE &$\checkmark$ &$\checkmark$ & 99.32 $\pm$ 1.97 & 19.43 $\pm$ 1.11  & 69.23 $\pm$ 0.66 & 18.02 $\pm$ 0.97 & 114.6  $\pm$ 0.65 & 20.24 $\pm$ 0.37 \\ 
    FedWeIT-MADE$^\ast$&$\checkmark$ &$\checkmark$ & 93.32 $\pm$ 2.70 & 14.43 $\pm$ 0.87  & 63.83 $\pm$ 1.12 & 12.40 $\pm$ 0.87 & 109.3 $\pm$  3.54 & 15.11 $\pm$ 0.71 \\ 
    CONFEDMADE & $\checkmark$ &  $\checkmark$& 87.12 $\pm$ 2.76 & 8.32 $\pm$ 0.76 & 59.15 $\pm$ 0.67& 8.12 $\pm$ 0.43 & 104.2 $\pm$ 3.10 &9.76 $\pm$ 0.87  \\
  \end{tabular}
  }
  \label{main_confed_table}
\end{table*}
\textbf{Learning settings and respective models.} We contrast CONFEDMADE with FedWeIT in the continual federated setting and relate it to offline scenarios and respectively FL, CL, and FCL baselines. Specifically, we consider \textbf{\textit{1) Single-Task Learning(Joint):}} In Offline, a standard MADE \citep{MADE} is optimized on the entire dataset (on all data) in a traditional training regime. Similarly, In \textbf{\textit{2) Federated Offline}}, all data is being evenly distributed across clients at the start, following the FedAvg MADE introduced in section 3.2 (with ablation study insights). \textbf{\textit{3) Continual Baselines:}} In CL-Finetune, \citep{hayeskanan} data is presented incrementally without revisits or employing any strategy for continual updates, expecting full catastrophic forgetting in a standard MADE model (with only 1 client). CL-Cumulative Replay (CR) \citep{hayeskanan} represents the best-case continual learning scenario, where data is introduced incrementally but accumulated over time (i.e. each already observed task remains always fully accessible), hence without incurring any forgetting, and finally EWC \citep{EWC}, a regularization based continual learning baseline. \textbf{\textit{4) CFL Baselines:}} We combine continual and federated scenarios, where the cumulative replay again accumulates all data individually per MADE client and finetune swaps out all data without. Again, in the spirit of bounds, the best model of section 3.2 is used. Additionally, we also compare against a few other FL baselines such as FedProx \citep{FedProx}, FedCurv \citep{FedCurv}, and a combination of FedProx and EWC evaluated in a continual setup. \textbf{\textit{5) FedWeIT-MADE and CONFEDMADE in CFL:}} the proposed unsupervised CFL approach (without data revisits) and divided into three models highlighting their gradual improvement a) a naive combination of FedWeIT and MADE, without insights of section 3.2 and naive application of eq. \ref{eq:fedweitloss}, b) FedWeIT-MADE$^{*}$, now including the ablation insights of section 3.2 to highlight their importance, c) our full CONFEDMADE approach, based on proposed eq. \ref{eq:Confed_loss}. We typically assume 5 clients and participation of all clients in each communication round in all FL (continual) scenarios. For simplicity, we train each task for 50 communication rounds, where each round is equivalent to one epoch. 

\subsection{Efficiently mitigating forgetting with CONFEDMADE across three data scenarios}
We present the final averaged log-likelihoods and forgetting values for all of our three sketched data sequences in experimental table \ref{main_confed_table}.
Starting with the full data learning scenarios, we can see that \textit{a single client (offline)} is only marginally better than the \textit{Cumulative Replay} scenario in federated and continual learning. Without constraints imposed on data seen by clients, it is understandable that these set the benchmark on achievable results with a MADE architecture. 

We continue to notice the non-surprising opposite trend with MADE in incremental and federated learning scenarios in their respective Finetune cases, where previously seen data samples are no longer continuously repeated with newly encountered data. Here, the frameworks naturally succumb to large catastrophic forgetting, as observed in the NLL-losses deteriorating to 115.31 and 126.12 compared to 74.47 and 73.32 respectively. Whereas, CL baseline like EWC only offers a small positive deviation from the opposite trend seen in the previous scenario. Upon introducing some form of continual learning notion into FL, we see an improvement in the model's capability to retain the old information with the \textit{baseline FCL techniques (FedProx, FedCurv, FedProx + EWC)}. \textit{FedCurv} stands out among the three techniques with an visible improvement of 10 nats in NLL loss. Overall, these approaches are limited as they either do not avoid negative interference or do so insufficiently by regularization alone. The naive FedWeIt-MADE already shows significant signs of improvement in the performance, highlighting our earlier intuition that masked autoencoders are effective when properly coupled with FedWeIt's proposed parameter decomposition in CFL scenarios (going beyond mere averaging and regularization).
Most importantly, upon enforcing the strategies discussed in section 3.3 (e.g. synchronized, communicated masking strategy) we see an additional improvement in the catastrophic forgetting as conceptualized in FedWeIT-MADE$^\ast$ . Finally, with our proposed CONFEDMADE approach performance over its two counterparts is further improved by roughly ~6.0 and ~10.0 nats. This can be attributed to the synergized masking strategy reducing parameter overlaps, order-agnostic training leading to an implicit model ensembling in training, and the augmented regularizers reducing the potential for interference when leveraging the knowledge base.
\begin{figure*}[t]
\includegraphics[width=0.33\linewidth]{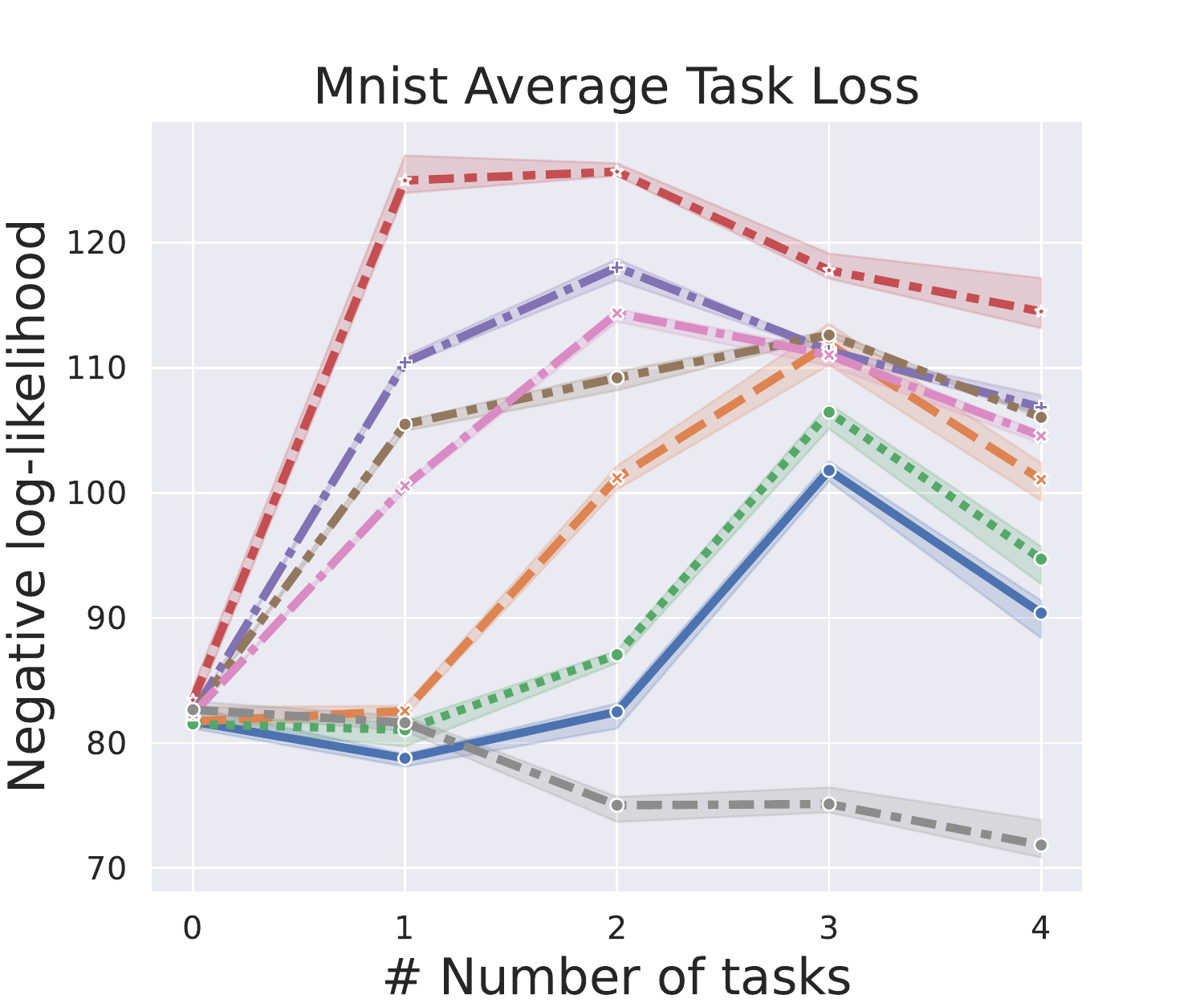}
   \hspace*{-24pt} \includegraphics[width=0.33\linewidth]{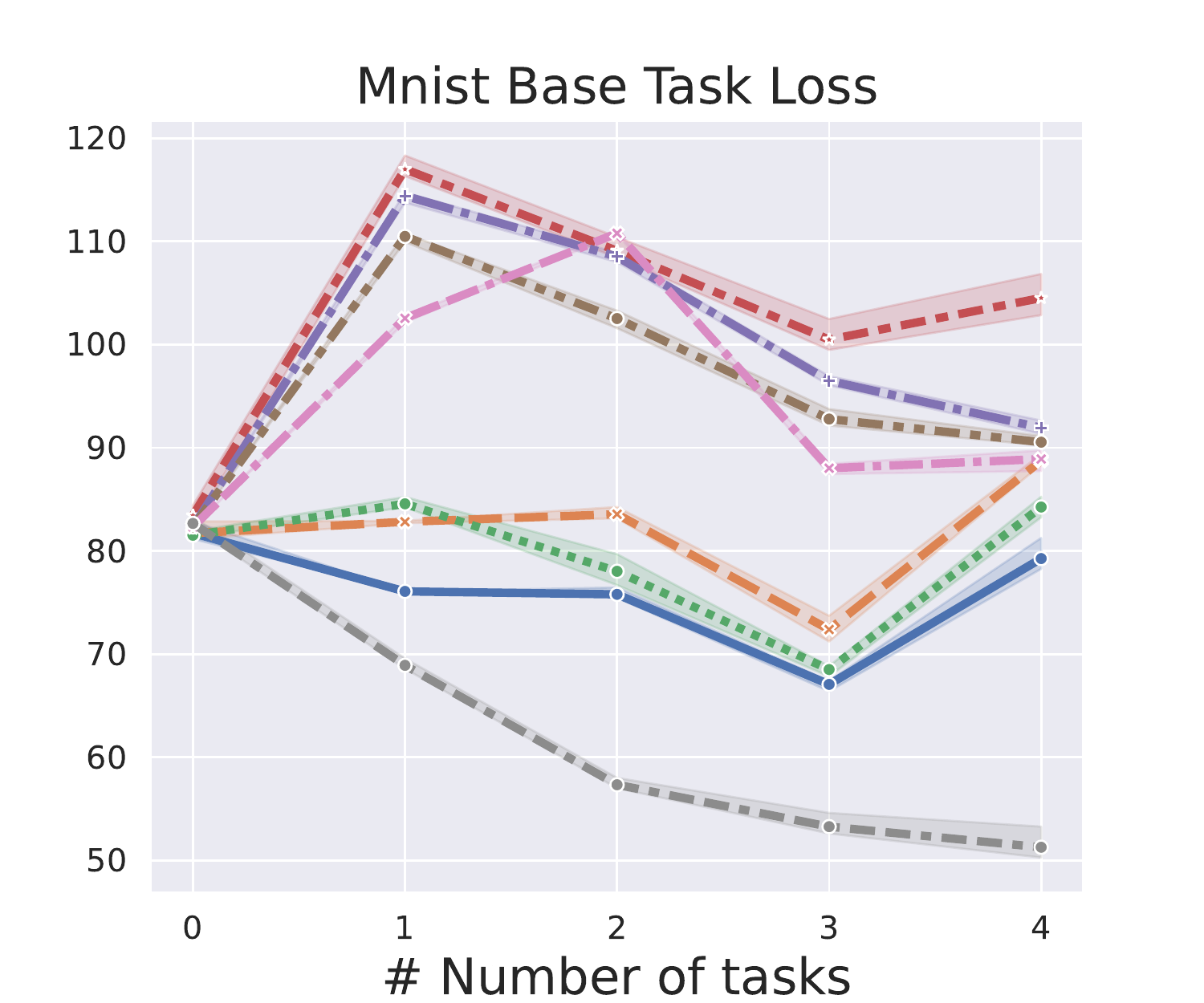} 
    \hspace*{-16pt}\includegraphics[width=0.385\linewidth]{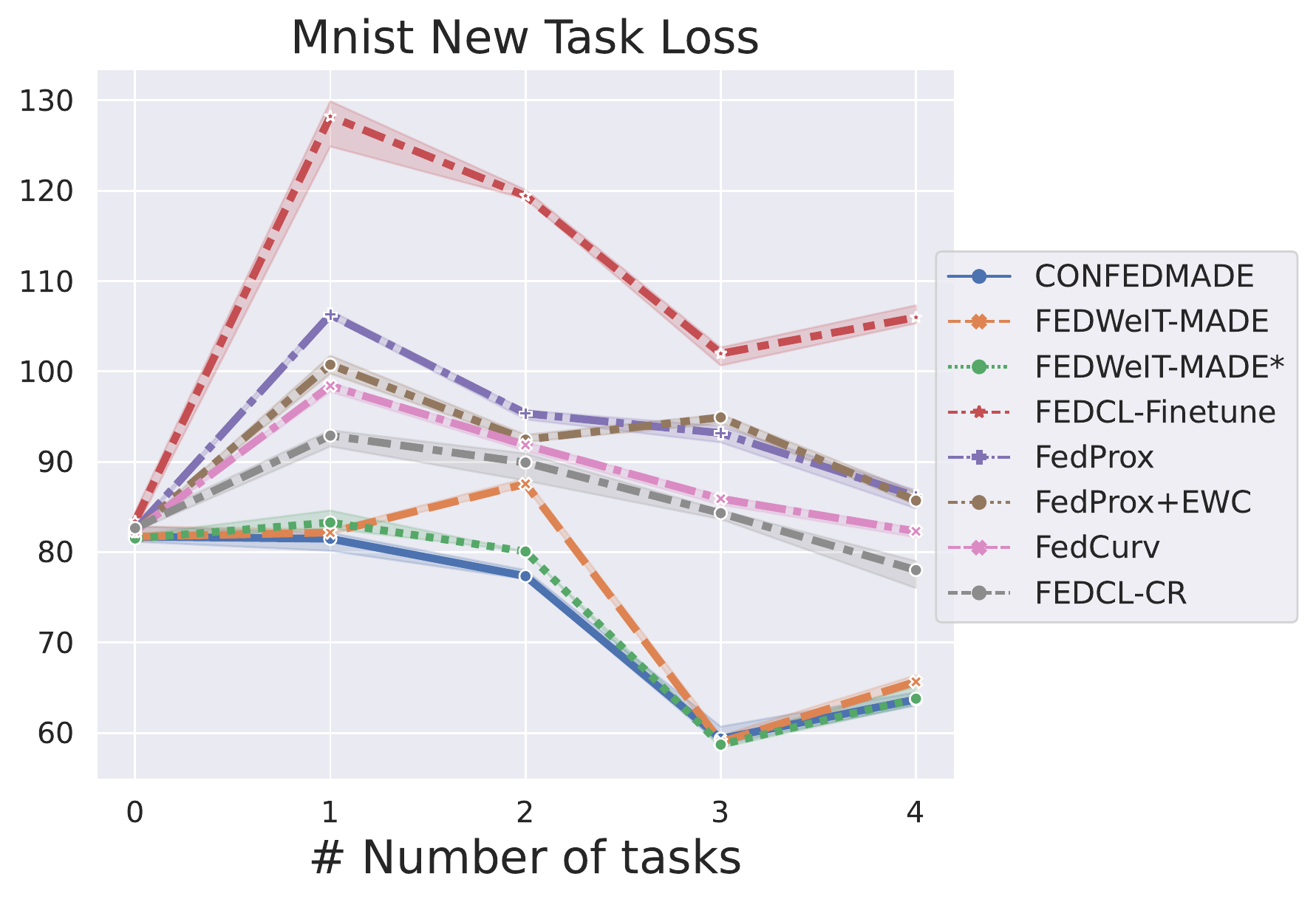}
    \caption{Decomposed negative log-likelihood in CFL to showcase: (left) an average of all the tasks seen so far, (center) the ``base'' task loss, i.e, the value for the only initial task in evolution over time to assess forgetting, (right) the ``new'' task loss, i.e., the value for only the newest task to gauge encoding of new knowledge. Lower values are better.}
    \label{Mnist}
\end{figure*}
To further nuance these statements, figure \ref{Mnist} shows the evolution of a decomposed loss after each task, separately quantifying average performance, forgetting the first task, and encoding new knowledge. We can observe that FedMADE forgets more of the learned representations of task 0 as we progress to learn. Although CONFEDMADE is yet to reach the full upper bound, it certainly reduces the gap to what can be achieved while widening the gap with a lower bound with a big difference of up to 20.00 in forgetting. Overall, the choices in section 3 (in the initial FedMADE$^\ast$) and the CONFEDMADE loss both meaningfully and significantly improve upon FedWeIT.

\textbf{CONFEDMADE attends to overlapping knowledge of other clients:}
To illustrate the effect of the inter-client transfer of learned representations, we plot the 'attention' values $\alpha$ at the end of the second task (one task on each axis) as heatmaps in figure \ref{heatmap attention}. The y-axis denotes the subset of the data seen by each client during the first task, whereas the x-axis denotes the data subset seen during the second task. In the figure's left panel, the tasks are set up in a way that there is a complete overlap of data subsets, whereas the right panel contains partial overlap among the clients.
 As such, we can investigate the strength of a CONFEDMADE client to decide when the shared knowledge across the network can aid its own training. Overall, we can observe higher values of $\alpha$ (0.60 and 0.67) in the respective plots when there is overlap in the data subsets (clients 1 and 4; clients 2 and 3) and thus later mitigated forgetting and values as low as 0.19 or 0.21 when there is no overlap of data subsets between the clients.

\textbf{CONFEDMADE reduces the cost of communication by actively masking model parameters:} As base parameters $B_c$ largely account for the communication cost of CONFEDMADE ($A_c$ is generally small as discussed in section 3.3), we investigate the sparsification/performance trade-off by ablating the auto-regressive MADE mask ($M^W$) and modulating the cut-off value for the FedWeIT mask ($m_c$) on MNIST in table \ref{table:n_clients}. Intuitively, we first quantify how much sparsification we can achieve at which cost through our inclusion of the auto-regressive masking. Here, we observe that we can fully leverage the advantages brought by FedWeIt's parameter decomposition framework, gaining a substantial, almost half, reduction in the communicated amount of $B_c$ at virtually no performance drop. In contrast, setting higher cut-offs on the FedWeIT mask (0-1 range), results in only a small further reduction of communication cost, with similar minor drops in performance. Despite this advantage seeming almost negligible in relation to the major gains obtained through MADE in CONFEDMADE, we can nevertheless observe that both masking strategies together are symbiotic. Ultimately, we have gained a continual federated unsupervised model with state-of-the-art likelihood estimates, while  \\
\begin{minipage}[]{\textwidth}
  \begin{minipage}[t!]{0.6\textwidth}
    \includegraphics[width=0.475\linewidth]{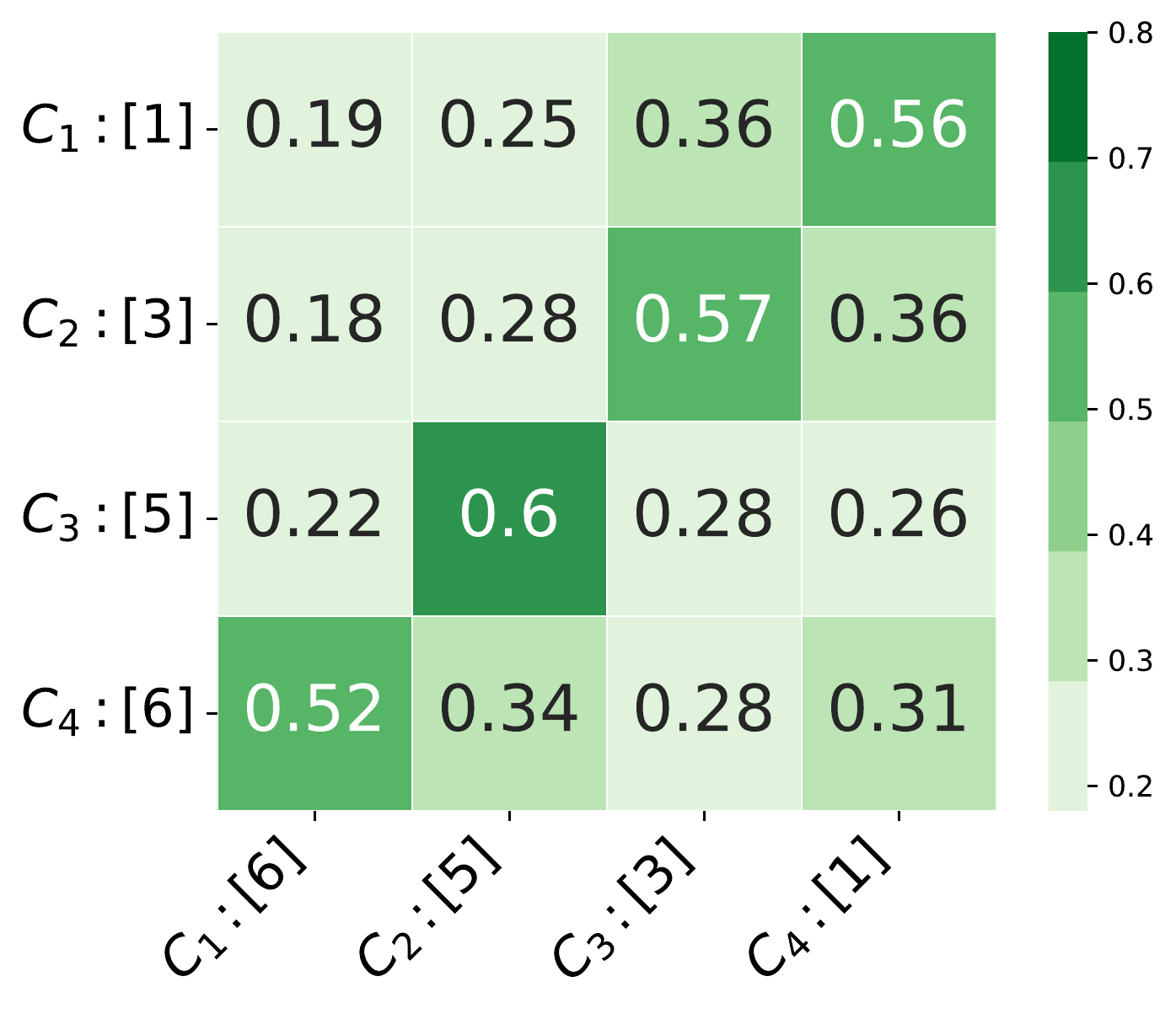}
    \includegraphics[width=0.485\linewidth]{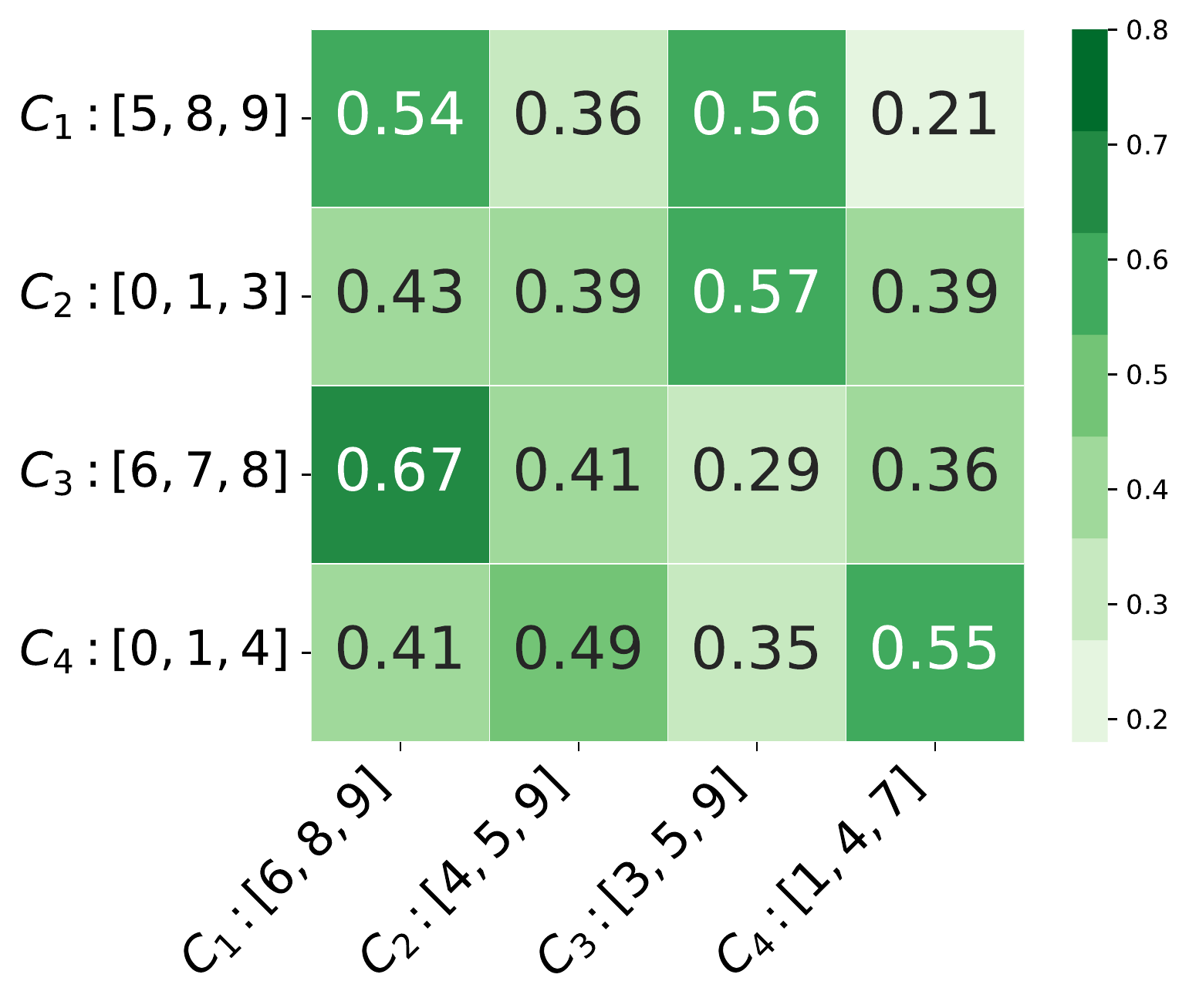}
    \captionof{figure}{Heatmaps for values of $\alpha$ (range 0 to 1) to highlight inter-client knowledge transfer when other clients have observed the same tasks (left) or have partial overlap (right). Two tasks for individual clients are denoted in brackets on x and y-axes respectively.}
    \label{heatmap attention}
  \end{minipage}
  \hfill
  \begin{minipage}[t!]{0.37\textwidth}
   \captionof{table}{Sparsification/Performance tradeoff to reduce communication through CONFEDMADE's $M^W$ and increased cut-off for FedWeIT $m_c$. Whereas $m_c$ by itself lightly improves communication at minor performance expense, CONFEDMADE lowers communication by almost 50\% at virtually no drop in NLL. \smallskip}
        \resizebox{\linewidth}{!}{%
        \begin{tabular}{@{}lcc@{}}
        \textbf{Comm. strategy} & \textbf{\% of $B_c$} & \textbf{NLL}  \\
         \toprule
        w/o $M^W$ \& $m_c>0.1$&  88.6  &  85.89 $\pm$ 0.67\\

        w/ $M^W$ \& $m_c>0.1$ &  44.5  & 87.12 $\pm$ 0.83\\
                \midrule
        w/ $M^W$ \& $m_c>0.2$  &  39.4  & 88.52 $\pm$ 0.44\\
        w/ $M^W$ \& $m_c>0.3$  &  34.7   & 90.08 $\pm$ 0.57 \\
        w/ $M^W$ \& $m_c>0.4$   &  30.0  & 91.12 $\pm$ 0.89\\
        \end{tabular}%
        }
        \smallskip
      \label{table:n_clients}
    \end{minipage}
\end{minipage} \\

leveraging FedWeIt's effective parameter decomposition and further reducing communication costs through MADE masking (we re-emphasize that MADE masking further allows to only communicate non-zero parameters on top of sparsification).

\section{Conclusion}
We have shown that masked autoencoders are effective continual federated learners by synergizing the benefits of auto-regressive masking with FedWeIT's parameter decomposition framework. Empirical evaluations on datasets ranging from image to numerical data have demonstrated that our introduced CONFEDMADE approach mitigates forgetting while substantially lowering communication costs. We believe such a finding will prove to be quite beneficial in a FL setup if the application in question becomes too complex and requires computationally expensive client architectures.
This work also paves the way for this relatively new field of research focusing on combining federated and continual setups for representation learning. 
One natural next step would be to scale this idea with the help of much more complex architectures such as Transformers where recent work has corroborated masked autoencoding as an effective strategy on a token basis \citep{MaskedTranformer}. The respective Transformer based approach can seamlessly be integrated with the auto-regressive masking strategies implemented in CONFEDMADE, allowing future investigation into clients that may additionally observe not only different distributions but also multiple modalities. 

\section*{Acknowledgements}
This work was supported by the project ``safeFBDC - Financial Big Data Cluster (FKZ: 01MK21002K)'', funded by the German Federal Ministry for Economics Affairs and Energy as part of the GAIA-x initiative. It benefited from the Hessian Ministry of Higher Education, Research, Science and the Arts (HMWK; project ``The Third Wave of AI''), the Federal Ministry of Educa- tion and Research (BMBF) and the State of Hesse Project collaborative center ``High-Performance Computing for Computational Engineering Sciences (NHR4CES)'' as part of the NHR Program. \\

\bibliography{references}
\bibliographystyle{references}
\newpage
\appendix
\section{Appendix}
In the appendix, we complement the main body with Training setup details and hyperparameters in part A and further empirical visualization in part B.
\section{Training Conditions and Hyperparameters}

\subsection{Training Configuration:} To evaluate our various experimental setups, we have used \textbf{\textit{NVIDIA TESLA V100 - SXM3}}. From the architectural point of view, a client model comprises of multi-layer perceptrons (MLP) with the varied hidden layer capacities depending on the complexity of the task to be learned. To put a constraint on the the model complexity, we have largely set the depth of the hidden layers between [350, 500] for images and [90, 110] for numerical data. We have also maintained a compression ratio roughly at 60\%. For the sake of simplicity, the number of clients typically ranges from 3 to 5 for our federated setups, whereas only a single client is used for continual setups. The duration of the experiments span from several hours to a max of 24 hours. We optimize our decomposed MADE client parameters using Adam or AdamW optimizers with the learning rate starting at $0.001$ alongside a weight decay of $0.0001$. We collaboratively learn individual task for 50 rounds of communication where each round resonates to only one epoch.  After extensive experimental evaluation, We have opted to use $0.0001$ and 100 for the hyperparameters $\lambda1$, $\lambda2$ respectively.  We have published our code at \url{https://github.com/ml-research/CONFEDMADE}.

\subsection{Task set Formulation}
Table \ref{tab:Dataset Table} provides a detailed description of the dataset as used in the main body. Recall, that it is implied that the offline scenario uses the entire dataset, whereas continual scenarios such as CL-Finetune and CL-Cumulative Replay make use of the respectively sketched tasks. The readers are reminded that the sequences of tasks are presented in isolation for the finetune and concatenated (accumulated) for the cumulative replay.

\subsection{Evaluation Measures}
In our main body's experiments, we have considered two sets of evaluation measures to compare our techniques. The first is based on the decomposition of the loss to measure the old and the new task \citep{kemker2017measuring}. The second is based on the notion of forgetting \citep{GEM}. 

\textbf{Average per-task Negative log-likelihood loss:} We measure the reconstruction loss at the end of the task t averaged over all the previous tasks. $N_t= 1/t \sum_{i=1}^t n_{t,i} $  where $n_{t,i}$ is the Negative log-likelihood (NLL) loss of task i after the completion of t.

\textbf{Base Task loss:} With this measure, we measure the ability of our frameworks to retain the information learned during the first task set at a future time step. Base task loss can be defined as $B_t = n_{t,0}$ where $n_{t,0}$ is the Negative log-likelihood of task 0 after the completion of task t. 

\textbf{New Task loss:} We also measure the model's capability to generalize on newly seen data samples. The New task loss for a client model can be defined as $N_t =  n_{t,t}$ where $n_{t_t}$ is the Negative log-likelihood for task t after the completion of task t. 

\textbf{Average forgetting:} We measure the forgetting in terms of the difference in the negative log-likelihood loss for task i after the completion of task t and its base task loss. For T tasks, Average forgetting can be defined as  $F_t= 1/(t-1) \sum_{i=1}^{t-1} max_{ t \epsilon 1.. T-1}(0,n_{t,i}-n_{T,i}) $  where $n_{t,i}$ is the negative log-likelihood for task i after the completion of task t.

\begin{table}[t]
\caption{We provide the details of the various dataset used in this work. The dataset ranges from Mnist, Emnist to Numerical datasets such as Adult, Connect4, etc. We have also stated total no of distinct subsets in each dataset (No. Classes), Sequence of tasks created (No. Task), number of distinct data subsets used per task (Classes per task). The train-validation-test splits can be visualized in the following Train, Validation, and Test columns.}
\label{dataset-table}

\setlength{\tabcolsep}{7pt}
\begin{center}

\resizebox{\linewidth}{!}{%
\begin{tabular}{lccccccc}
\toprule
\multicolumn{5}{c}{\textbf{DATASETS}} \\
\toprule
\textbf{Dataset }           & \textbf{No. Task} & \textbf{No. Classes}  & \textbf{Classes per task} & \textbf{Train} & \textbf{Validation}  & \textbf{Test}  \\
\toprule
\textbf{Mnist} \citep{MNi} & 5 & 10 & 3 &  15620 &   3121  &  3120 \\
\midrule
\textbf{Mnist} \citep{MNi} & 5 & 10  & 1 &  5626 &   1126  &  1127 \\
\textbf{Emnist} \citep{emnist} & 5 & 10 & 13 &  42817 &   8564  &  8563 \\
\midrule
\textbf{Adult} \citep{UCI} & 4 & 1 & 1 &  23257 &   4653  &  4651 \\
\textbf{Connect4} \citep{UCI} & 4 & 1 & 1 &  48255 &   9651  &  9651 \\

\textbf{Tretail} \citep{UCI} & 4 & 1 & 1 &  20990 &   4199  &  4198 \\
\textbf{RCV1} \citep{UCI} & 4 & 1 & 1 &  34285 &   6858  &  6857 \\

\bottomrule
\end{tabular}
}

\end{center}
\label{tab:Dataset Table}
\end{table} 


\subsection{Description of the working of the mask}
We briefly describe the strategies and fundamentals to formulate the auto-regressive masks utilized in the form of the masked autoencoders. To be precise, referring to the main script, the left figure in Fig 2 presents the structure of a masked autoencoder where the shaded lines represents the masked connections and the solid lines are the connected ones. In the scope of our, the input sample is assumed to be of D dimensions, and for the sake of simplicity, we have chosen it to be 3. The input layer as a result is seen to be numbered accordingly as 1, 2, 3, which in a way also resembles its order. Being an autoencoder, the output layer also consists of the similar number of units as the input layer. For all the hidden layers in the network consisting of K hidden units, we sample an integer denoted by m(k) for each unit such that it is between 1 and D-1 (the numbers inside the circles). MADE adheres to the autoregressive property, so a unit $x_d$ only depends on its previous input units $x_{<d}$. To reinstate such a principle, the integer m(k) assigned to a unit is compared against all the units in the previous layer. Thereby, if m(k) for a unit is less than the value of the unit (number inside the circle) in the previous layer, we mask that weighted connection between them (shown in grey shaded lines). This rule remains valid for all the connections between the input to hidden and hidden to hidden layers. For the connection between the hidden and the output layer, we mask the weighted connection if m(k) of a hidden unit is greater than the value assigned to the output unit. We follow these rules to mask only the weighted connections between the layers.

Meanwhile in case of Order-agnostic connectivity, at every mini-batch or epoch, we shuffle the orderings of the input nodes at layer 0. By doing so, we need to repeat the process of determining the connections to be masked in relation to the values assigned in the next layers. In connectivity-agnostic scenarios, we rather the re-sample the values m(k) assigned to every nodes of the respective hidden layers.

\begin{figure}[]
    \centering
    \, \includegraphics[width=0.325\linewidth]{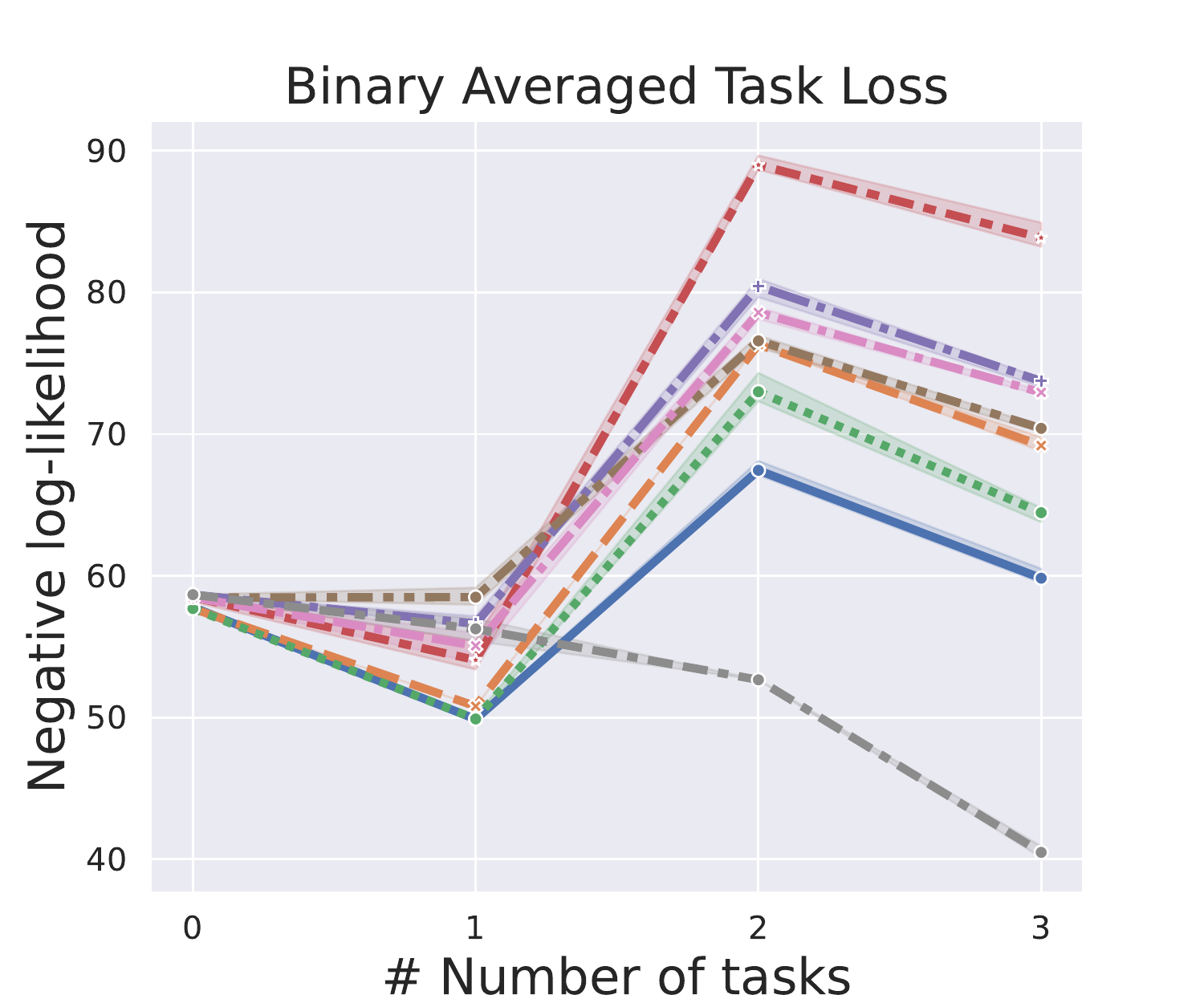} 
    \includegraphics[width=0.325\linewidth]{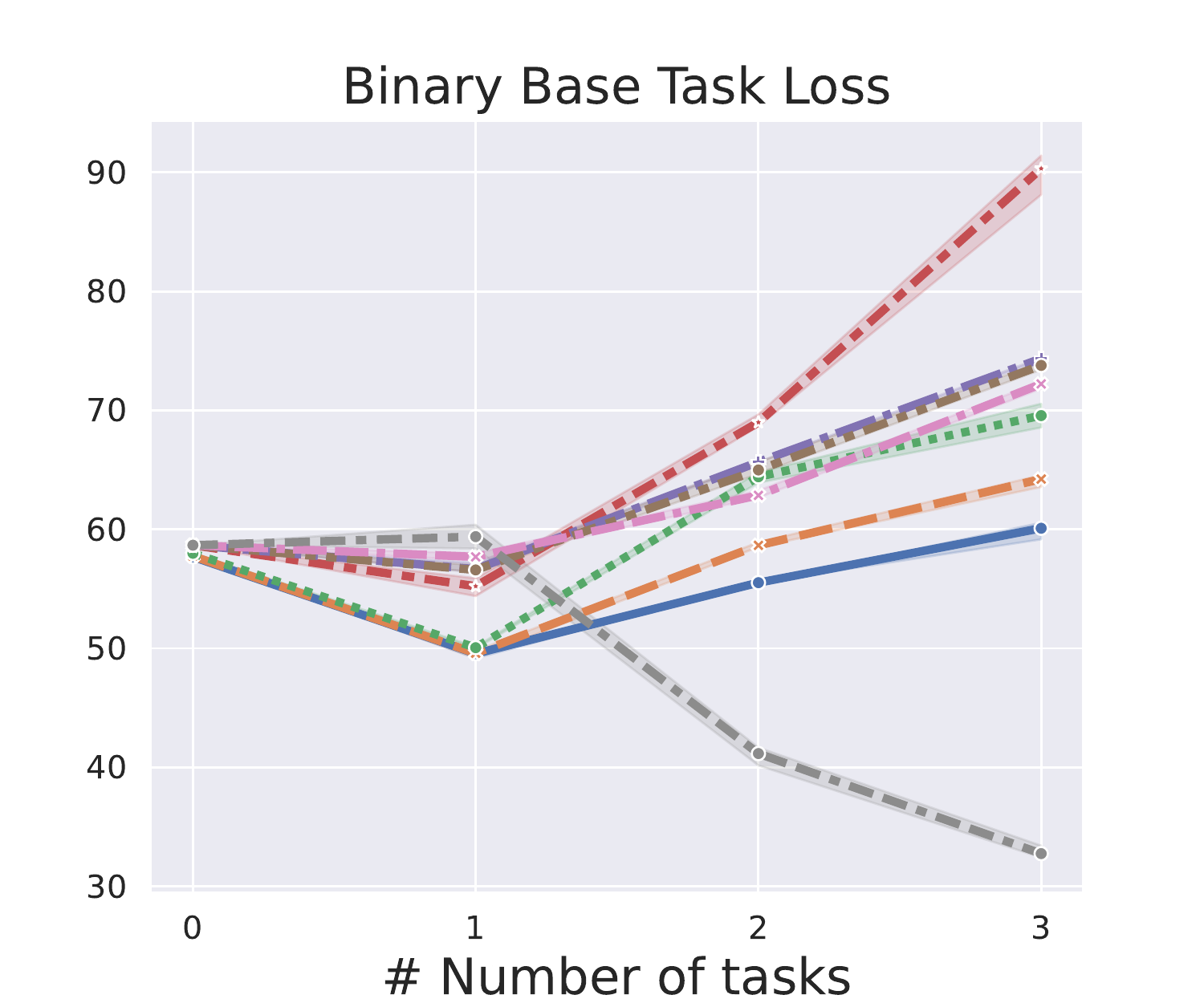}
    \includegraphics[width=0.325\linewidth]{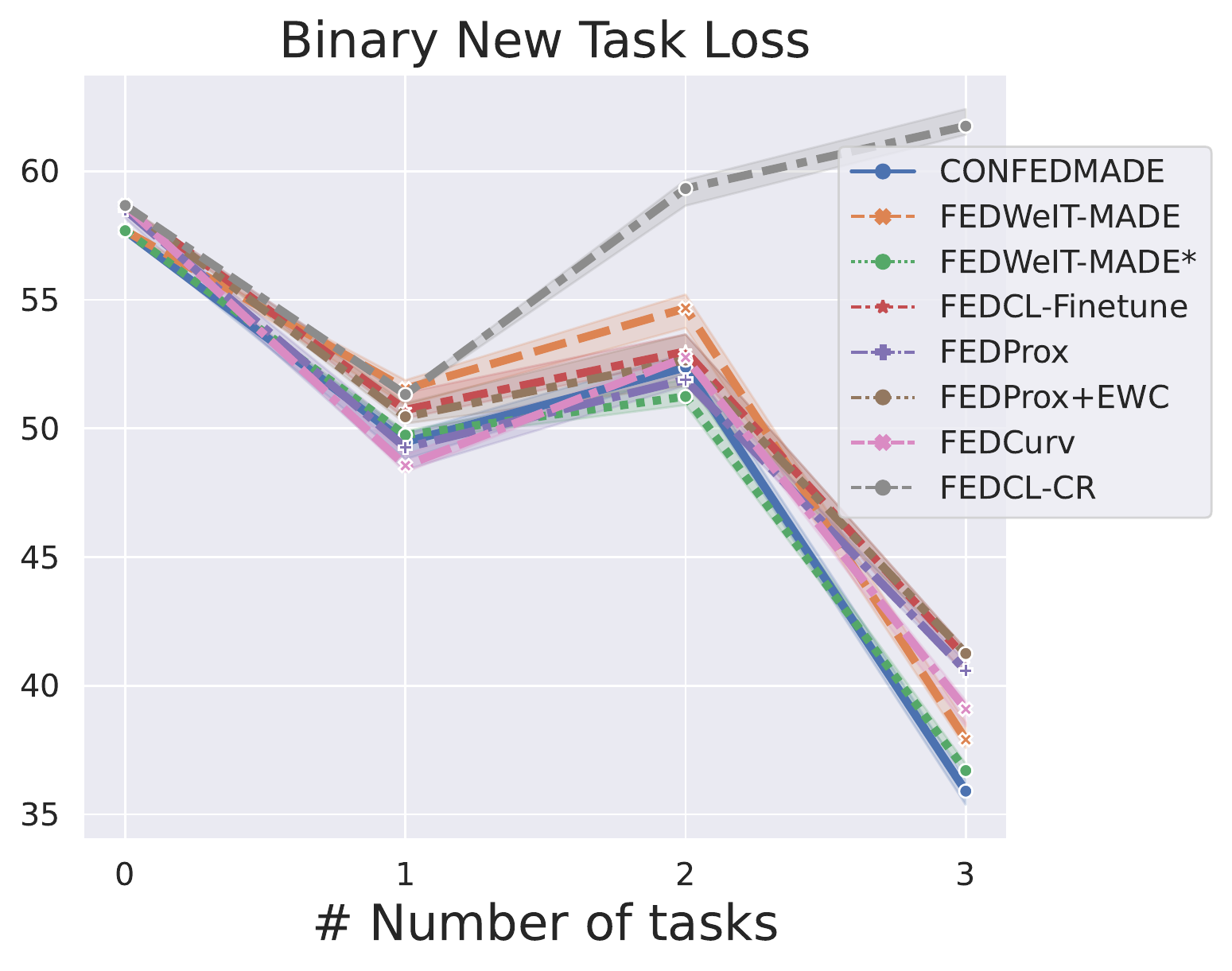}
    \caption{Decomposed negative log-likelihood in FCL for Binary datasets to showcase: (left) average of tasks seen so far, (center) the ``base'' loss, i.e the value for only initial task in evolution over time to assess forgetting, (right) the ``new'' loss, i.e. the value for only the newest task to gauge encoding of new knowledge. Lower values are better.} 
    \label{tab: Mnist}  
\end{figure}
\subsection{Empirical Visualization}
Recall in the main body, we have visualized the effectiveness of CONFEDMADE against other approaches for Sequential Mnist dataset \citep{MNi}. To further validate that we can scale our proposed approach against different data distributions, we produce the three similar plots for Binary datasets based on Average task loss, Base task loss and New task loss. If we refer to the first plot on the left (Average task loss), CONFEDMADE can be seen to retain previously learned representations much more efficiently compared to its other counterparts under similar continual lower-bound scenario. Thereby, we can establish CONFEDMADE as effective continual federated learner, pertaining to our continual and auto-regressive masking strategies.

\end{document}